\documentclass[10pt,twocolumn,letterpaper]{article}

\usepackage{cvpr}
\usepackage{times}
\usepackage{epsfig}
\usepackage{graphicx}
\usepackage{amsmath}
\usepackage{amssymb}
\usepackage{caption}
\usepackage{authblk}
\usepackage{multirow}
\usepackage{booktabs}

\usepackage[pagebackref=true,breaklinks=true,letterpaper=true,colorlinks,bookmarks=false]{hyperref}

\cvprfinalcopy 

\begin{document}


\title{Sparsely Grouped Multi-task Generative Adversarial Networks for Facial Attribute Manipulation}

\author{Jichao Zhang$^1$,~Yezhi Shu$^1$,~Songhua Xu$^2$,~Gongze Cao$^3$,~Fan Zhong$^1$,~Meng Liu$^1$,~Xueying Qin$^1$}

\affil{%
    $^1$Shandong University, %
    $^2$Xi'an Jiaotong University, %
    $^3$Zhejiang University
}

\makeatother

\maketitle

\begin{abstract}
Recent Image-to-Image Translation algorithms have achieved significant progress in neural style transfer and image attribute manipulation tasks. However, existing approaches require exhaustively labelling training data, which is labor demanding, difficult to scale up, and hard to migrate into new domains. To overcome such a key limitation, we propose Sparsely Grouped Generative Adversarial Networks (SG-GAN) as a novel approach that can translate images on sparsely grouped datasets where only a few samples for training are labelled. Using a novel one-input multi-output architecture, SG-GAN is well-suited for tackling sparsely grouped learning and multi-task learning. The proposed model can translate images among multiple groups
using only a single commonly trained model. To experimentally validate advantages of the new model, we apply the proposed method to tackle a series of attribute
manipulation tasks for facial images. Experimental results demonstrate that SG-GAN can generate image translation results of comparable quality with baselines methods on adequately
labelled datasets and results of superior quality on sparsely grouped datasets. The official implementation is publicly available~\footnote{Code: https://github.com/zhangqianhui/Sparsely-Grouped-GAN.}.
\end{abstract}

\section{Introduction}
Learning the mapping from a source image to a target image has rich
applications in computer vision and graphics. For example, the problem of image style transfer, which transfers the style
of one image to that of another image while keeping the content of the second image unchanged, can be regarded as an image-to-image learning task.
Previous style-transfer methods~\cite{DBLP:journals/corr/GatysEB15a, Li:2017:LNS:3123266.3123425}
take advantage of the powerful representation capability of deep convolutional neural networks and use Gram matrices of
neural activations to represent the artistic style of an image. However, its iterative optimization process for stylizing an image consumes much time. Besides leveraging pre-trained deep models for deriving style features, a recent method~\cite{isola2017image} based on Generative Adversarial Networks(GAN)~\cite{goodfellow2014generative}
learns an end-to-end mapping from the style of an input image to that of another image. Guided by the adversarial loss, the discriminator can flexibly
learn a similarity measurement optimized for a specific task instead of relying upon a hand-engineering one, the learning capability of which enables the network to generate samples with a high visual quality. Besides image style transfer, Isola $et.al.$ ~\cite{isola2017image} applies Generative Adversarial Networks for other kinds of image-to-image learning tasks, such as translating color
images to edge maps, facial attribute manipulation, and labels to street scenes. Their work aims to learn a common model for tackling multiple image-to-image mapping tasks, the task of which is also commonly referred to as an Image-to-Image Translation~(IIT) task.

Previous IIT methods can be categorized into two broad classes, including
\emph{supervised learning from paired training data} and \emph{unsupervised learning from grouped~(unpaired) training data}.
The first method that learn from paired data requires that each source image in a training set should be explicitly associated with a corresponding target image. Collecting such training data requires non-trivial labelling efforts, in particular when a dataset is sizable. To alleviate this burden in gathering supervised training data, the second method learns from grouped training data instead. These methods accept training datasets where a group of source images is associated with another group of target images. In this way, there is no longer a need to specify the one-to-one correspondence between the two groups of images; only group-level correspondence information is anticipated, which significantly reduces the amount of annotation efforts. Nevertheless, when the quantity of groups needed and training samples for every group is sizeable, it still can be laborious to supply the group correspondence information required by the second class of methods.

\begin{figure}[t]
\begin{center}
\includegraphics[width=1.0\linewidth]{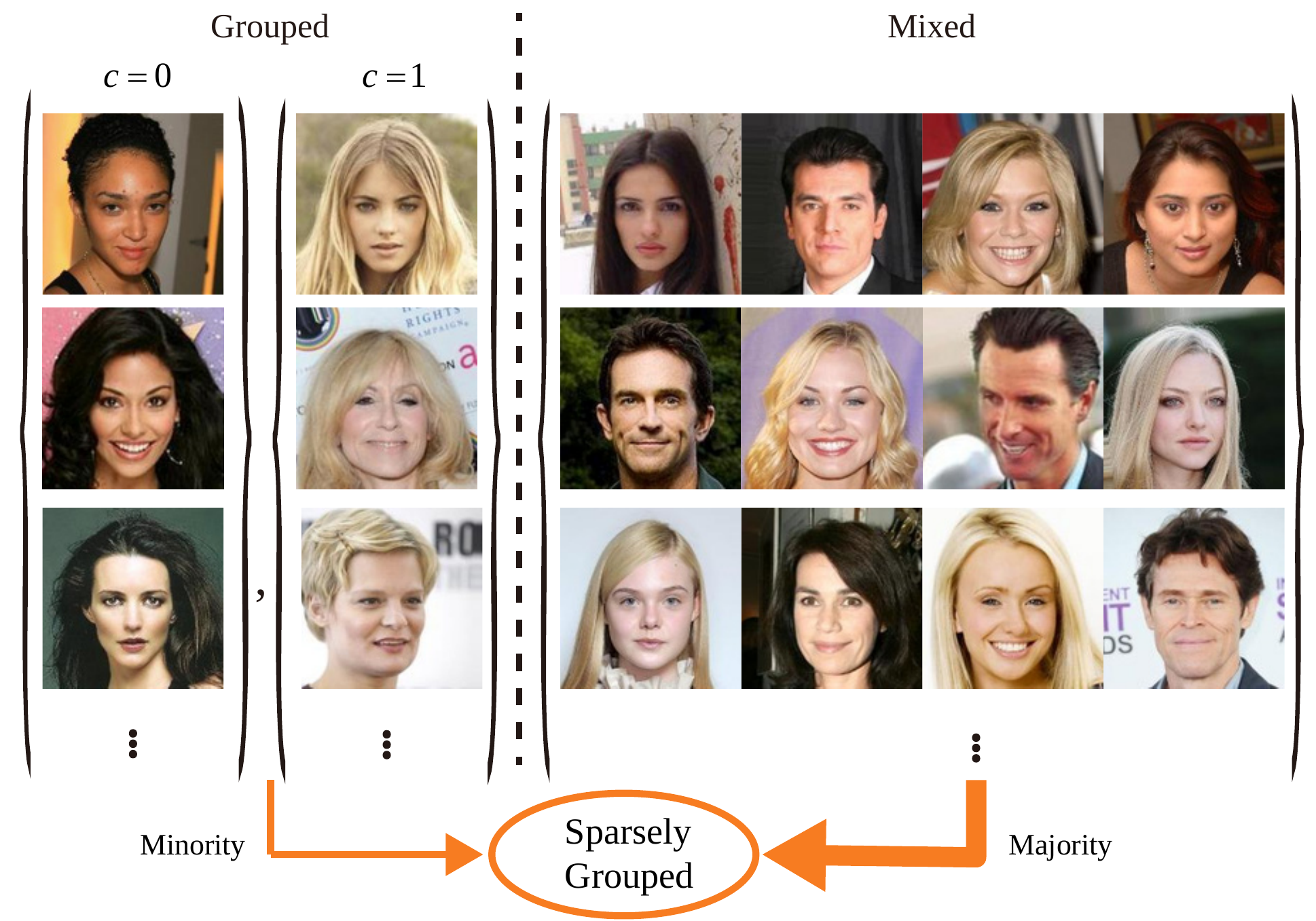}
\end{center}
\captionsetup{font={small}}
\vspace{-0.4cm}
\caption{Introduction of sparsely grouped dataset. A sparsely grouped dataset consists of a minority of grouped-labelled data and a majority of mixed data.
Here, grouped training dataset consists of two groups, one for black hair with $c=0$ and the other for blond hair with $c=1$.}
\label{fig:domain}
\end{figure}

Recognizing the above limitations of the existing methods, the key motivation of this work is to perform image-to-image translation involving multiple groups using only a single model.
Most of existing IIT models learn the translation between two domains. For example, if we want to translate a male facial image to a corresponding female facial image, to translate a facial image with black hair color into the corresponding facial image with the blown hair color, or from an image of a
young person to the image of the person in senior ages. In general, these existing methods require a dedicated IIT model to be separately trained for every attribute. Such an approach can thus be inefficient and ineffective in conducting multi-group image translation tasks. To address this
problem, StarGAN~\cite{choi2017stargan} was proposed, which is an unified generative adversarial network capable of learning the mapping
among multiple groups with a high visual quality as demonstrated through experiments carried out on various datasets in comparison with prior models. However, StarGAN requires multi-attribute
labels in the target domain and leverages a single generation model, whose image translation result for one attribute can be undesirably affected by the manipulation over other attributes. This problem becomes more pronounced when the model is trained on unbalanced datasets.

To overcome the above problems, we propose a novel sparsely grouped learning method, where only a few of the training dataset
is grouped while the remaining unlabelled data~(i.e. mixed data shown in Fig.~\ref{fig:domain}) is used for unsupervised learning
to improve the performance of classification and stabilize the training of the adversarial network. To execute an IIT task, the proposed model only needs a few training samples per group, thus greatly reducing human labeling efforts in data preparation. To the best of our knowledge, no existing image
translation architecture or off-the-shelf tool can be readily and directly applied to learn from sparsely grouped datasets. To address the gap, we propose a
one-input multiple-output network, called \emph{Sparsely Grouped Generative Adversarial Networks(SG-GAN)}, for learning to translate the image
attribute between a paired group. Moreover, our architecture can be extended to perform IIT for multiple groups with multi-task learning. Different
from StarGAN with the target domain as the input of generator, SG-GAN would define all output domains with different network branch which can reduce the negative impact on a given translation task from other parallel translation tasks.

Overall, the main contributions of this work include:
\begin{itemize}
\item A novel model, SG-GAN, is proposed for tackling image-to-image translation tasks by learning the mapping among multiple groups on a sparsely grouped dataset where only a small portion of data points are annotated with group-level labels and the group affiliation information for the rest of data points remains unknown.
\item We further devise a refined residual image learning component into the proposed SG-GAN to improve the degree of translation
for the target image attribute involved in a translation process while preserving other visual attributes and background unrelated to the translation goal in the generation results.
\item The proposed model can generate comparable facial attribute translation results using much fewer grouped-labelled samples than peer methods.
\end{itemize}

\section{Related Work}

\begin{figure*}[t]
\begin{center}
\includegraphics[width=1.0\linewidth]{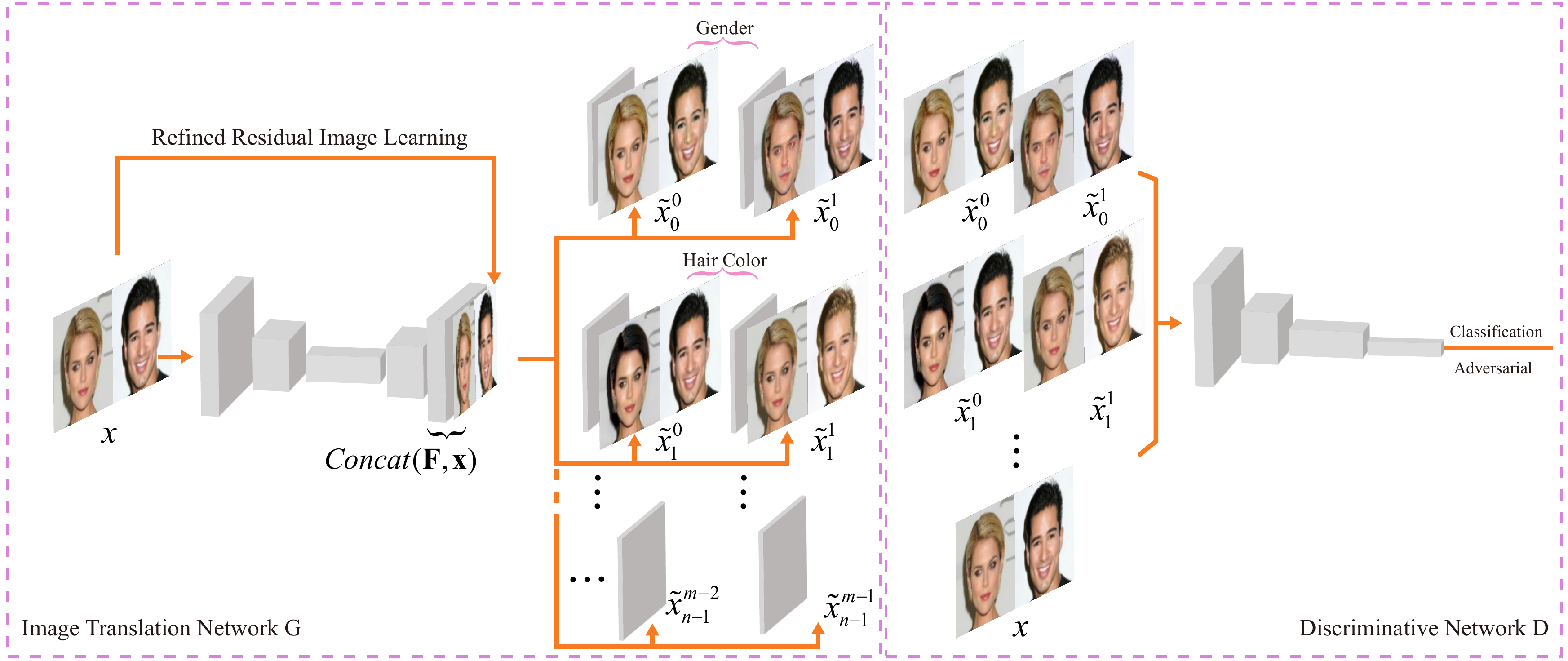}
\end{center}
\vspace{-0.4cm}
\caption{Multi-task learning architecture of SG-GAN. $\mathbf{G}$ is a one-input, multi-output network that performs multiple
attributes manipulation(e.g., gender(male/female), hair color(black/blond)). Refined residual image learning just concatenates
input $\mathbf{x}$ into the middle feature maps $\mathbf{F}$ of decode network to get $Concat(\mathbf{F},\mathbf{x})$ which is
as the input of the final convolution layer. Note that $\mathbf{D}$ will be a semi-supervised classifier when trained in the sparsely grouped dataset.}
\label{fig:model}
\end{figure*}

{\bfseries Generative Adversarial Networks:}
Generative Adversarial Networks(GAN)~\cite{goodfellow2014generative} is a powerful implicit generative
model to produce a model distribution that mimics a given target distribution,
 and it has been applied to many fields, for example, low-level image processing
 tasks~(image in-painting~\cite{pathak2016context, IizukaSIGGRAPH2017}, image super-resolution~\cite{Ledig_2017_CVPR}),
 high-level semantic or style transfer~\cite{li2016precomputed, Zhao:2017:SAA:3123266.3123450, DBLP:journals/corr/GatysEB15a, huang2018munit, DRIT, Siarohin_2018_CVPR, 8636265},
 video prediction and generation~\cite{mathieu2015deep, Pan:2017:CYT:3123266.3127905, wang2018pix2pixHD}).


Modelling the high-dimensional distribution of data and generating high-solution, photo-realistic images are an important
research spot for generative models. Recently, Karras \emph{et al.}~\cite{karras2017progressive}
proposed a new training methodology for GAN, which can progressively train the network's generator and discriminator.
Their method can generate highly realistic facial images of $1024^{2}$ pixels in a completely unsupervised manner. Guo \emph{et al.}~\cite{DBLP:journals/corr/abs-1903-11250} devise a novel Auto-Embedding Generative Adversarial Networks~(AEGAN) that generates high resolution images~($512^{2}$ pixels) by learning a latent embedding extacted from an autoencoder.
For supervised learning for the conditional generation, StackGAN~\cite{Zhang_2017_ICCV} for text-image synthesizing
could generate realistic $256^{2}$ images conditioned on only text descriptions. It is very difficult to
generate high-quality images when training complex datasets, such as ImageNet~\cite{5206848}. Recently, DeepMind~\cite{anonymous2019large}
has proposed a new class-conditional image synthesis models, called BigGAN, successfully generating high-resolution, diverse samples with ImageNet
dataset as the training dataset, and greatly improves the assessment results in FID~\cite{heusel2017gans} and Inception scores~\cite{salimans2016improved}.
In general, the powerful modelling ability of GAN also provides a substantial basis for the wide range of applications in other fields.

Though GAN is a very powerful generative model, one of the key challenges of
GAN is the instability of its training because of the vanishing gradient.
 The objective functions of GAN are carefully designed to attain more stable training. For example, LS-GAN~\cite{Mao_2017_ICCV}
adopts a least squared loss function in its discriminator to solve the vanishing gradient problem.
The Wasserstein GAN(WGAN) proposed in~\cite{pmlr-v70-arjovsky17a} uses
the Wasserstein distance instead of the Jensen-Shannon distance to form its objective function,
which achieves a more stable training process. For vanilla GAN, Miyato \emph{et al.}~\cite{DBLP:journals/corr/abs-1802-05957}
proposes a novel weight normalization, called spectral normalization, to stabilize the training of networks by
limiting the spectral norm of the weight matrices for the discriminator. These techniques
have been applied for IIT tasks to get higher-quality translation results.

{\bfseries Image-to-Image translation(IIT):}

The essence of IIT is to learn the mapping between pairs or groups of images while preserving image characteristics irrelevant
to the current translation task. The previous work of IIT using conditional GAN~\cite{isola2017image} has attained impressive results.
Their method applies a supervised learning-based approach on pairs of images in the training phase, which incurs a major limitation
for large-scale application and migration into new domains and tasks. To overcome the limitation mentioned above, a collection of IIT methods based on learning from grouped-labelled training data were proposed,
e.g.,~\cite{Yin:2017:LGE:3123266.3123423,dong2017unsupervised,Zhu_2017_ICCV, pmlr-v70-kim17a, shen2017learning,
DBLP:journals/corr/abs-1709-06548},
to reduce the ground truth paired-labels efforts for training data preparation.

In addition to directly learning the mapping from one or multiple source images to one or multiple corresponding target images,
another thread of active research endeavors to conduct disentangled representation learning, the result of which can then be
applied for facial attribution manipulation in images~\cite{larsen2015autoencoding, DBLP:journals/corr/DumoulinBPLAMC16,
brock2016neural, NeuralFace2017, xiao2018elegant, lample2017fader, he2017arbitrary},
image style-transfer~\cite{DRIT, huang2018munit, liu2018unified}. In general, these models
would learn to encode the input image into two spaces: 1) a domain-invariance content space
and 2) a domain-specific attribute space. To learn the disentangled representation,
the grouped-labels will be direct or indirect as the input of GAN for supervised learning.

The research trends of IIT tasks can be summarized in three aspects: 1) translation from a single image to video sequence and 2)
translation from low-solution sample to high-solution sample and 3) translation from one-task-one-model to multi-task-one-model. In details,
Pix2PixHD~\cite{wang2018pix2pixHD} has been proposed for synthesizing 1024$\times$2048 photo-realistic images from semantic label maps
using conditional generative adversarial networks and the similar idea, called vid2vid~\cite{wang2018vid2vid} has been applied for video-to-video translation task.
Recently, StarGAN~\cite{choi2017stargan} was proposed, which utilizes a GAN-based architecture to learn a series of mappings from a common group of
source images to multiple groups of target images using a single generative model. In comparison, our SG-GAN algorithm proposed in this paper
requires a much smaller amount of ground truth grouped-labels in its learning process, thanks to its semi-supervised learning framework. Moreover,
StarGAN would suffer the severe problems, where the translation results of the single target attribute are easy to be affected by other attributes,
especially when trained on unbalanced samples between the negative attribute value and positive attribute value. Our SG-GAN solves this problem with the elaborate network designing.

{\bfseries Facial Attribute Manipulation:}

Facial attribute manipulation is a IIT task, which aims at manipulating the semantic content of a facial image according
to a specified attribute value. \cite{larsen2015autoencoding} proposed a novel model by combining a variational autoencoder with a generative adversarial networks, for facial attribute manipulation but acquires labelled data to compute the visual attribute vectors in the testing. Zhang~\emph{et al.}~\cite{pmlr-v77-zhang17c} proposed a model called ST-GAN, which could be trained on a
 mixed dataset to establish relationships between latent codes and generated samples for semantic information discovery. Even though the design of their model is novel and heuristic, ST-GAN lacks of high-quality manipulation results.

Recently, GAN-based residual image learning has been applied to facial attribute manipulation, the method of which
is referred to ResidualGAN~\cite{shen2017learning}. ResidualGAN attains satisfactory translation results. In comparison,
the proposed SG-GAN can obtain multiple facial attribute manipulation effects using only a single trained model, which also
produces more visually realistic IIT results. Motivated by human attention mechanism
theories~\cite{rensink2000dynamic}, attention mechanism has been successfully introduced in image classification~\cite{DBLP:journals/corr/XiaoXYZPZ14},
image segmentation~\cite{DBLP:journals/corr/ChenYWXY15} and natural language processing task~\cite{xu2016ask}. Recently, the attention model
has been applied for the facial attribute manipulation field by learning the attention maps of face to alter the attribute-specific region and keep
the rest unchanged~\cite{pumarola2018ganimation,Zhang_2018_ECCV}.

{\bfseries Semi-Supervised Learning using GAN:}

Sparsely grouped learning tasks can be regarded as a type of semi-supervised learning tasks. With a large amount of labelled data,
deep neural networks have achieved great development in many application areas such as computer vision and natural language processing; yet it has been a challenge to apply deep models to the dataset with limited labels. So a substantial amount of efforts has been
dedicated to addressing semi-supervised learning tasks. The CatGANs~\cite{DBLP:journals/corr/Springenberg15} have proposed categorical generative adversarial networks, a new model for robust unsupervised learning and semi-supervised learning. For semi-supervised with GAN, the discriminator would also be the classifier for predicting class distribution from unlabelled
and labelled samples. It is reasonable that this should work. By adversarial learning with unlabelled samples, the presentations learned
by discriminator help to improve the classification, where the structure present by learning from unlabelled data contains information
that can be used to infer the unknown labels. In addition to
classification tasks, ~\cite{ DBLP:journals/corr/SoulySS17} propose a method for semi-supervised
semantic segmentation using adversarial networks and~\cite{2016arXiv160507725M} achieves state-of-art performance for multiple semi-supervised
text classification tasks, including sentiment classification and topic classification.
The control of the number of labelled data can alleviate the problem of the limited dataset mentioned before. In our method,
using only a minority of grouped-labelled data for translation learning, we can also achieve high visual quality image translation result.

\section{Methods}

\subsection{Generative Adversarial Networks}

Goodfellow \emph{et al.}~\cite{goodfellow2014generative} propose the generative adversarial networks~(GANs) which is a powerful implicit generative model for modeling the complex, high-dimensional distributions of the images. The objective function of GANs is given as follows:
\begin{eqnarray}
\mathop{min}\limits_{G}\mathop{max}\limits_{D}\ell(D,G) &=& \mathbb{\mathbb{E}}_{\mathbf{x}}[logD(\mathbf{x})] \nonumber \\
&+& \mathbb{\mathbb{E}}_{\mathbf{z}}[log(1-D(G(\mathbf{z})))]
\label{eq_gan_loss}
\end{eqnarray}

\subsection{SG-GAN}

{\bfseries One-Input Multiple-Output Architecture for Multi-task Learning:}

Unlike the vanilla GAN model~\cite{goodfellow2014generative}, which directly learns the mapping from a noise vector $\mathbf{z}$ to an image $\mathbf{x}$, $G$ in IIT task learns the mapping from an input image $\mathbf{x}$ to an output image $\mathbf{\tilde x}$, where it can be regarded as an ``autoencoder"~\cite{article}.

Some previous methods have two generator networks with input images from different groups~\cite{choi2017stargan,Zhu_2017_ICCV}.
However, such architecture, which consists of two generators with input data from different groups, could not work well in the sparsely grouped dataset where most data
are mixed while only a few data are with grouped-labels. It is hard to learn a precise mapping from a source group to a target group using few group-labelled
samples. Recently, StarGAN~\cite{choi2017stargan} uses a single generator network for manipulating multiple attributes.
It is difficult to train their original model for sparsely grouped learning because of the lack of original domain labels to perform the reconstruction loss for training data without grouped-labels. To address the gap, we propose a one-input multiple-output  network for learning to translate the image
attribute between a paired group and would introduce the details of this model below.

Given a face image $\mathbf{x}$ with n attributes $\mathbf{C}=\{\mathbf{c}_{1}, \mathbf{c}_{2}, \mathbf{c}_{3}, ..., \mathbf{c}_{n}\}$,
where the value range of each $\mathbf{c}_{i}$ is $m$. As illustrated in Fig.~\ref{fig:model}, we propose a
one-input multi-output architecture as generator $G$, which maps the input image $\mathbf{x}$ to all existing groups.
Supposed that $n=2, m=2$, we would use $G$ to map the face image $\mathbf{x}$ with original group
$\mathbf{C}=\{\mathbf{c}_{1}=[0,1], \mathbf{c}_{2}=[0,1]\}$ into four target groups, which
are $\mathbf{\tilde C^{0}_{0}}=\{\mathbf{\tilde c}_{1}=[0,1], \mathbf{\tilde c}_{2}=[0,1]\}$, $\mathbf{\tilde C^{1}_{0}}=\{\mathbf{\tilde c}_{1}=[1,0], \mathbf{\tilde c}_{2}=[0,1]\}$,
$\mathbf{\tilde C^{0}_{1}}=\{\mathbf{\tilde c}_{1}=[0,1], \mathbf{\tilde c}_{2}=[1,0]\}$, $\mathbf{\tilde C^{1}_{1}}=\{\mathbf{\tilde c}_{1}=[1,0], \mathbf{\tilde c}_{2}=[1,0]\}$, respectively. For input sample $x$ without grouped-label $\mathbf{C}$, it would also be translated into these four groups. In the case of attribute gender, we would map an input face $\mathbf{x}$ into a male and a female result regardless of their original gender. Such a design of the generator network can work independently of original group-labels of the input data.
It is evident that both labelled and unlabelled face images can be used as input to $G$.
The final output of $G$ is $\mathbf{\tilde x}^i_{j}, 0 \leq i < m, 0 \leq j < n$, where they would be feed into the discriminator.
$D$ is also a multi-task learning network, which performs adversarial-learning and
classified-learning with multiple facial attributes. In addition to adversarial learning,
$D$ network can be regarded as multiple classifiers with outputting $n$ logit vectors where every vector with $\mathbf{m}$-dim is
used as input to a softmax function. The output dimensions of $G$ and $D$
will increase as the number of facial attributes participating in the manipulation grows.
Similar to StarGAN and CycleGAN, the architecture of our generator $G$ can be regarded as  ``autoencoder".
But the difference is that the final layers of our model do not share variables, where
every translation has a separate network. To some extent, we demonstrate
experimentally that this new setup could reduce the negative impact on the
current attribute translation from others when training the model for multiple tasks learning.

{\bfseries Sparsely Grouped Learning:}
Different from performing the adversarial loss and classification loss with a single network to
classify $(m+1)$ class, where the generated samples are classified into
the $\left(m+1\right)$-th class, we output the adversarial logits and classification logits using
two separate networks in the final layer. We construct a discriminative network $D$
for classification and adversarial learning to distinguish generated samples $\mathbf{\tilde x}$ and real
samples $\mathbf{x}$. $D$ for the classified-learning needs to classify
input samples by the attribute value and builds the objective function for every attribute.
The network training for every attribute has the same objective function.
Given the real image $\mathbf{x}$ with the original group $\mathbf{c}_{j}$ for the $j$-th facial
attribute, the objective function of $D$ for this attribute is computed as the softmax loss:
\begin{equation}
\ell_{d\_cls} = \mathbb{\mathbb{E}}_{\mathbf{x}, \mathbf{c}_{j}}[-log(D(\mathbf{c}_{j} | \mathbf{x})],
\end{equation}
where $D(\mathbf{c}_{j} | \mathbf{x})$ is the softmax probability over this group label $\mathbf{c}_{j}$.
It is noted that this classification loss is just for the grouped samples with labels.
In the same way, for the generated samples $\mathbf{\tilde x}_{j}$ with the target group $\mathbf{\tilde c}_{j}$,
the softmax objective function for training $G$ is:
\begin{equation}
\ell_{g\_cls} = \mathbb{\mathbb{E}}_{\mathbf{\tilde x},
\mathbf{\tilde c}_{j}}[-log(D(\mathbf{\tilde c}_{j} | \mathbf{\tilde x}_{j})]
\end{equation}

Different from the previous GAN model, SG-GAN would output multiple generated samples $\mathbf{\tilde x}^{i}_{j}$, $i=0,1,...,m-1$ for the $j$-th
facial attribute. The min-max adversarial loss for generator $G$ and discriminator $D$ with this facial attribute is:

\begin{equation} \label{eq_final_gan_loss}
\begin{aligned}
\ell_{adv} = \mathbb{E}_{\mathbf{x}}[logD(\mathbf{x})] + \sum_{i=0}^{\mathbf{m}-1}
[\mathbb{E}_{\mathbf{\tilde x}^i_{j}}[log(1-D(\mathbf{\tilde x}^i_{j}))]]
\end{aligned}
\end{equation}

{\bfseries Refined Residual Image Learning:}

The residual image learning was proposed by~\cite{shen2017learning} aims to
improve the effectiveness of facial attribute manipulation and make the modest modification of the attribute-specific facial area
while keeping irrelevant attribute or content unchanged. However, this method with the residual image learning, which sums the natural images
and outputs of the network directly is hard to generate very realistic images, especially based on the vanilla GAN loss~\cite{goodfellow2014generative}.
We also found that the residual image learned tends to be sparse and empty when using very powerful
GAN loss, e.g., the GAN loss from WGAN-GP~\cite{gulrajani2017improved}.

To alleviate these problems, we improve the previous residual learning method in two aspects. As shown in Fig.~\ref{fig:model}, one is to use the concatenation
of feature maps $\mathbf{F}$ and input $\mathbf{x}$ instead of the element-wise sum to automatically learn the relation of arithmetic
 between the feature maps $\mathbf{F}$ and the input $\mathbf{x}$, the other is that we add the extra convolution layer to
refine the concatenation results $Concat(\mathbf{F}, \mathbf{x})$ to get the final translation results. We name
this method with tiny changes as the refined residual image learning and adapt it to our architecture.

{\bfseries Reconstruction Loss:}
By minimizing the adversarial and classification losses, $G$ is trained to generate very realistic samples and translate the input
into the correct target group. However, minimizing these losses does not guarantee that translation results preserve the identity
and background content of the input images. So, similar to the goal of refined residual image learning with some modifications in architecture,
we hope to alleviate this problem by adding a new loss function. The previous methods~\cite{Zhu_2017_ICCV, choi2017stargan} apply a cycle consistency loss to the generator. This cycle consistency loss is proposed to measure the discrepancy occurred when the translated image is brought back to the original image space. However, this loss does not adapt to our architecture.
We propose a reconstruction loss for the generator $G$. For the single attribute, this loss can be defined as

\begin{equation} \label{recon_loss}
\begin{aligned}
\ell_{rec} &=& \mathbb{E}_{\mathbf{x}}[\Vert G(G(\mathbf{x})^i)^j - G(\mathbf{x})^j \Vert_{1}
\\ &+& \Vert G(G(\mathbf{x})^j)^i - G(\mathbf{x})^i \Vert_{1}],
\end{aligned}
\end{equation}

where $G(\mathbf{x})^i$ and $G(\mathbf{x})^j$ mean the $i$-th and $j$-th output of $G$ network~($i \ne j$).
Supposed that $i=0,j=1$, $G$ network uses $\mathbf{x}$ as input to obtain the translation result $\mathbf{\tilde x}^{0}$ and $\mathbf{\tilde x}^{1}$.
Then, these result will also be new input to obtain a new translation $\mathbf{\tilde X}^{1}$ and $\mathbf{\tilde X}^{0}$ of corresponding groups.
We adopt this $L1$ norm as the loss and minimize it to make $\mathbf{\tilde x}^{0}$ and $\mathbf{\tilde X}^{0}$, $\mathbf{\tilde x}^{1}$ and $\mathbf{\tilde X}^{1}$ as close as possible.

{\bfseries Overall Objective Function:}
The objective functions for discriminator $D$ will be different for the mixed data and grouped data, as $\ell_{\mathbf{d\_cls}}$ is just for the grouped data
with labels. Finally, the full objective functions for the specific facial attribute to optimize model $D$ is shown as:

\begin{eqnarray}
\ell_{D} =
\begin{cases}
\ell_{d\_cls} - \ell_{adv} &\text{Grouped},\\

- \ell_{adv} &

\text{Mixed}.
\end{cases}
\end{eqnarray}

The objective function for generator $G$ with this facial attribute is:
\begin{equation}
\ell_{G} = \ell_{g\_cls} + \lambda_{adv} \ell_{adv} + \lambda_{rec} \ell_{rec},
\end{equation}
where $\lambda_{\mathbf{adv}}$ and $\lambda_{\mathbf{rec}}$ is constant weight for adversarial loss and reconstruction loss.
We use $\lambda_{\mathbf{adv}}=1$ and $\lambda_{\mathbf{rec}}=10$ in all our experiments.

{\bfseries Network Architecture:} Our architecture is similar to previous IIT methods, which have shown impressive results for style transfer and semantic manipulation. Our generator $G$ contains convolution layers with the stride size of two for encoding, some residual blocks for expanding the receptive field, transposed convolution layers for decoding. We use instance normalization~\cite{articlein} for the generator but no normalization for the discriminator. Note that $tanh$ activation function is used for output of the generator. More details about network architecture are shown in the Appendix~\ref{net_arch}.

{\bfseries Training Details:}
We apply new technique from Wasserstein GAN with gradient penalty~(WGAN-GP)~\cite{gulrajani2017improved} to stabilize training process and generate high quality images. We replace Eq.~\ref{eq_final_gan_loss} with the new object function of WGAN-GP defined as:
\begin{eqnarray}
\ell_{adv} = \mathbb{E}_{x}[D(x)] - \sum_{i=0}^{m-1}(\mathbb{E}_{\tilde x^i_{j}}[D(\tilde x^i_{j})] - \lambda \mathbb E(t^{i}_{j})),
\end{eqnarray}
where $t^{i}_{j}$ is a gradient penalty variable for training $D$ and more details can be found in~\cite{gulrajani2017improved}. $\lambda$ is 10 for all our experiments.

We use the Adam~\cite{DBLP:journals/corr/KingmaB14} with $\beta_{1}=0.5$ and $\beta_{2}=0.999$. The size of the training batch is set to 8 for our experiments. Similar to WGAN-GP~\cite{gulrajani2017improved}, the generator is updated once after every five updates performed over the discriminator. Our all models are trained with the learning rate of 0.0001 for the first 10000 iterations and the learning rate will be linearly decayed to 0 over the next 10000 iterations.

{\bfseries Coping with Unbalanced Dataset:} SG-GAN starts with an undersampling procedure to balance the majority and minority groups. However, rather than discarding data through an undersampling process by existing methods, the proposed method uses a follow-up semi-supervised learning procedure where unsampled data elements are observed as unlabelled records during a learning process. In this way, all dataset elements are effectively utilized for model training, without overlooking informational clues carried by any one of them.

\begin{figure*}[t]
\vspace{-0.4cm}
\begin{center}
\includegraphics[width=1.0\linewidth]{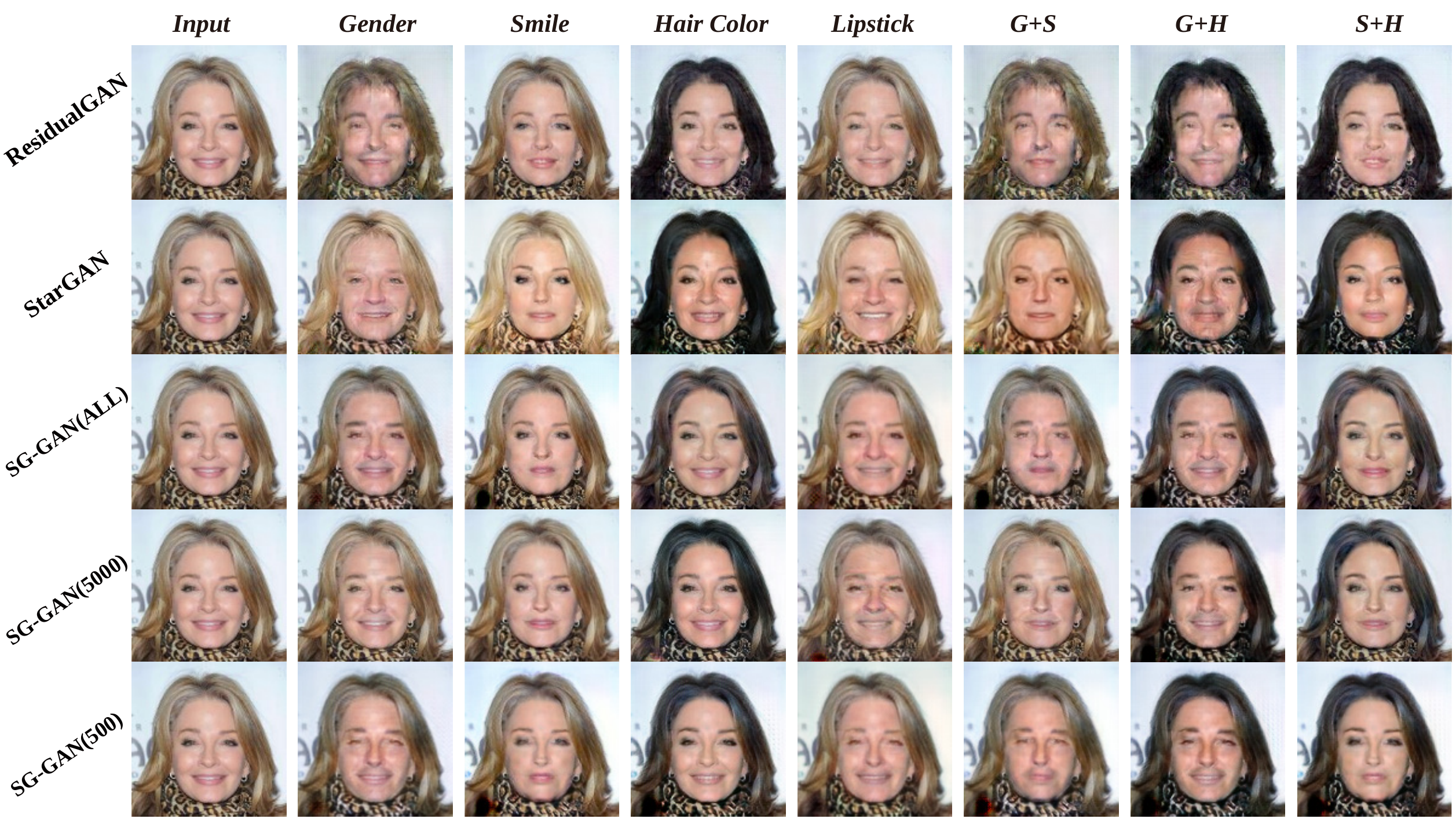}
\end{center}
\vspace{-0.4cm}
\caption{Comparison of facial attribute manipulation results on CelebA test dataset. The 1st and 2nd rows show facial attributes manipulation results of the baseline methods, i.e. ResidualGAN and StarGAN; the 3rd row shows the results of the proposed model in the condition that all
data are grouped; the results are shown in the last two rows when using 5000 and 500 images for every value of the attribute as the grouped data,
respectively. G: gender; S: smile; H: hair color.}
\label{fig:comparison}
\vspace{-0.2cm}
\end{figure*}

\begin{figure*}[t]
\begin{center}
\includegraphics[width=1.0\linewidth]{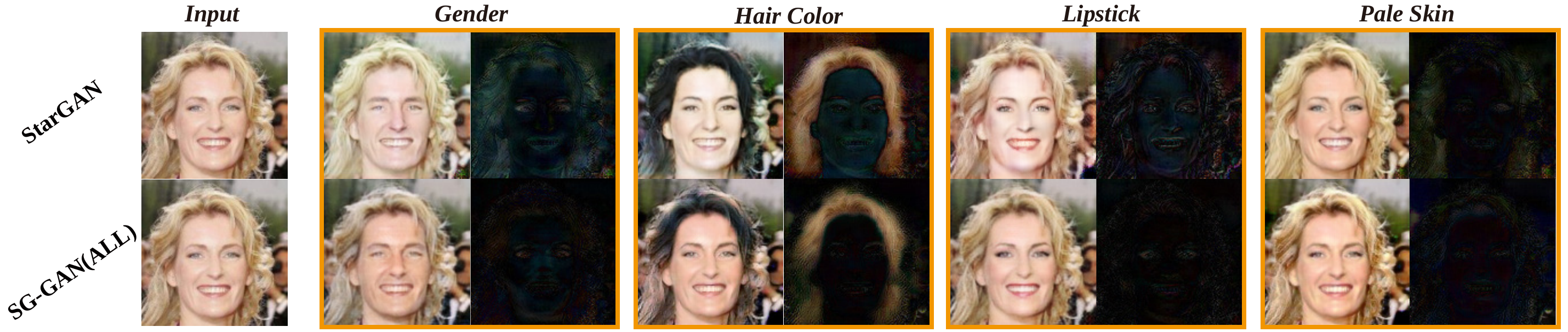}
\end{center}
\vspace{-0.4cm}
\caption{Manipulation results on targeted attribute affected by other unrelated attributes.
StarGAN and SG-GAN(ALL) both are trained on grouped dataset with four attributes: gender,
hair color, lipstick, pale skin. In addition to translation results for every attribute, the absolute
value of the residual image between input images and translation results are shown on the right.}
\label{fig:figure_archi_dis}
\vspace{-0.5cm}
\end{figure*}


\section{Experiments}
We firstly compare SG-GAN against recent methods of facial attribute manipulation tasks via both qualitative and
quantitative evaluations. Next, we analyze some problems of the previous network architecture
and demonstrate that the validity and superiority of our architecture. Lastly, we conduct an ablation study and compare
the proposed method against several reduced variants to validate the effectiveness of the refined residual
image learning component in our method.
\subsection{Baseline Models}

\begin{figure*}[!htp]
\begin{center}
\includegraphics[width=1.0\linewidth]{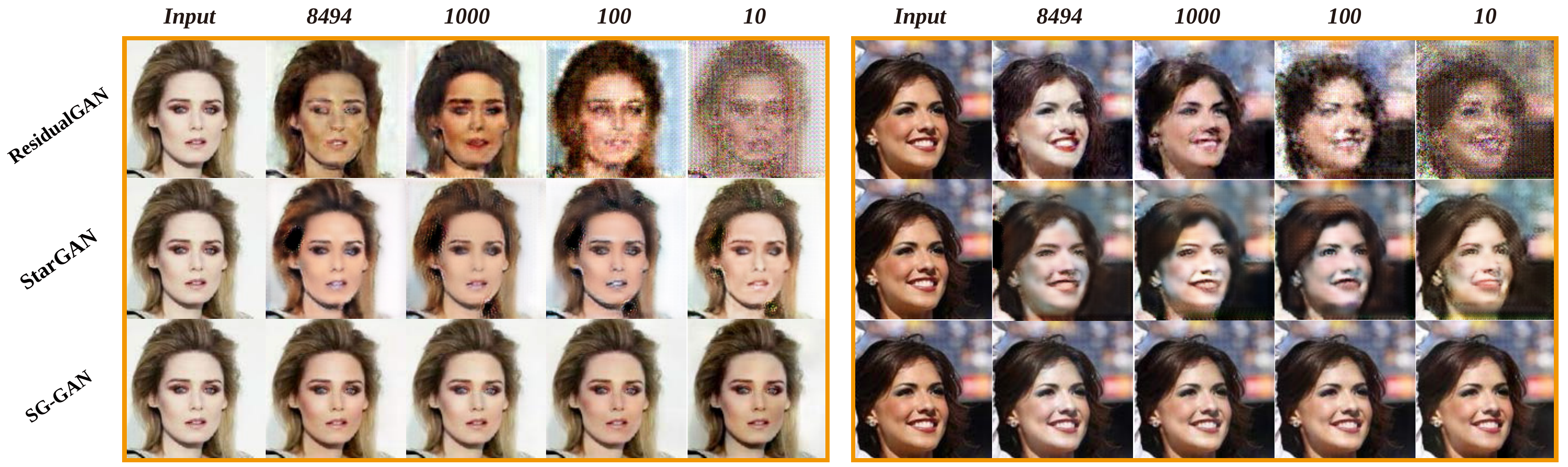}
\end{center}
\vspace{-0.4cm}
\caption{Translation results of the facial attribute pale skin on the different sparsely grouped dataset.
The three rows show results using different models: ResidualGAN, StarGAN and SG-GAN,
respectively. In the left, the 1st column images show input images while the next
columns are generated results using the different number of data with grouped-labels. The comparison results
for another person have been shown in the right.}
\label{fig:comparison_unb}
\end{figure*}

\begin{table*}
\small
\caption{Inception-scores for different models on a single attribute transfer task.}
\vspace{-0.4cm}
\begin{center}
\begin{tabular}{ccccc}
\hline
Attribute & Gender & Smile & Hair Color & Lipstick \\
\hline
ResidualGAN & 3.24$\pm$0.118 & 3.67$\pm$0.130& 2.88$\pm$0.196 & 3.49$\pm$0.193 \\
StarGAN & 3.36$\pm$0.164 & 3.62$\pm$0.104& {\bfseries 3.44$\pm$0.110} & 3.33$\pm$0.150 \\
SG-GAN(ALL) & {\bfseries 3.53$\pm$0.220} &  {\bfseries 3.68$\pm$0.135} & 3.23$\pm$0.174 & {\bfseries 3.50$\pm$0.116} \\
SG-GAN(5000) & 3.50$\pm$0.160 & 3.61$\pm$0.104 & 3.20$\pm$0.152 & 3.48$\pm$0.136 \\
SG-GAN(500) & 3.30$\pm$0.147 & 3.49$\pm$0.145 &  3.11$\pm$0.23 & 3.36$\pm$0.104 \\
\hline
CelebA   & \multicolumn{4}{c}{3.83$\pm$0.167} \\
\end{tabular}
\end{center}
\label{tab:inception}
\vspace{-0.2cm}
\end{table*}

\begin{table*}
\small
\caption{Classification accuracy [\%] of transfer results on the attribute gender and hair color for different models.}
\vspace{-0.3cm}
\begin{center}
\begin{tabular}{cccccccccc}
\hline
\multirow{2}{*}{Attribute} & \multicolumn{4}{c}{Gender} & \multicolumn{4}{c}{Hair Color} \\
\cmidrule(r){2-5}
\cmidrule(r){6-9}
& Gender & Smile & Hair Color & Background
& Gender & Smile & Hair Color & Background \\
\hline
ResidualGAN & 95.49 & 85.80 & 95.90 & 63 & 83.15 & 86.07 & 93.41 & 18 & \\
StarGAN & 96.23  &  85.99 & {\bfseries 99.63} & 64 & {\bfseries 99.43} & 88.26 & {\bfseries 99.53} & 63 & \\
SG-GAN(ALL) & {\bfseries 99.03} &  86.87 & 97.99 & {\bfseries 72} & 91.38 &  87.42 & 98.99 & {\bfseries 73}  \\
SG-GAN(5000) & 93.79 & {\bfseries 87.34} & 98.42 & 71 & 89.58 & 87.36 & 95.72 & 72 & \\
SG-GAN(500) & 93.36 & 84.87 & 97.90 & 70 & 89.11 & {\bfseries 88.68} & 88.91 & 69 & \\
\hline
CelebA & 99.00 & 90.22 & 99.53 & 100 & 99.00 & 90.22 & 99.53 & 100 & \\
\end{tabular}
\end{center}
\label{tab:gender_classification}
\vspace{-0.2cm}
\end{table*}

{\bfseries Multiple facial attributes manipulation:}
For multiple facial attributes manipulation, StarGAN~\cite{choi2017stargan} has achieved more high-quality facial attribute manipulation results
compared with DIAT~\cite{DBLP:journals/corr/LiZZ16e}, ICGAN~\cite{Perarnau2016}, CycleGAN~\cite{Zhu_2017_ICCV}. So, we just compare our model against StarGAN model.

We also adopt ResidualGAN~\cite{shen2017learning} as a baseline which performs facial attributes transfer work with the residual image learning.
Both StarGAN and ResidualGAN belong to~\emph{method that learn from grouped training data} and they require all training data to be labelled for different groups.
 More importantly, ResidualGAN acquires the equal number of images between the positive value and the negative value for the highly unbalanced attribute in the training process. Unlike ResidualGAN,
 StarGAN just uses all labelled data, does not take some measures to solve the unbalance problem of the training data. Additionally, for multi-attribute manipulation, we must train ResidualGAN many times.

We implement the code of ResidualGAN by ourselves and use the official code of StarGAN\footnote{Please see https://github.com/yunjey/StarGAN}.

\subsection{Dataset}
{\bfseries CelebA Dataset:} The CelebFaces Attributes Dataset (CelebA)~\cite{liu2015deep} contains 202,599 face images
with large pose variations and background clutter. Each of image has annotation of 40 binary attributes. We crop and scale the images to 128$\times$128 pixels. We randomly
select 5,000 images as the test dataset and use the remaining images as the training dataset. In our experiments, we
select multiple facial attributes, e.g., gender, hair color~(black hair and blond hair), smile for manipulation.

\subsection{Baseline Comparison on CelebA dataset}
In this section, we provide comparison results with SG-GAN and baseline methods in some binary attribute manipulation   tasks and use some metrics to evaluate the translation results from multiple different perspectives.

{\bfseries Qualitative evaluation:}
Fig.~\ref{fig:comparison} shows facial attribute manipulation results generated by SG-GAN with three sparsely grouped datasets,
which respectively have all samples, 5000 samples and 500 samples with grouped-labels for every value in the attribute,
where the sparse rates for these three sparsely grouped datasets are 1, 0.05 and 0.005, respectively.
To distinguish these models, we denote them as SG-GAN(ALL), SG-GAN(5000), SG-GAN(500) respectively.
Note that, we would repeat these data with grouped-labels many times to increase the probability of their participation in training for every iteration.
As shown in the 3rd to the 5th rows of Fig.~\ref{fig:comparison}, the quality and degree of translation results does not noticeably
decline when the number of data with grouped-labelled images is mostly reduced. These experimental results show
SG-GAN still could learn an exceptional binary classifier, which is the basis of learning perfect image translation when the data with group-labelled is scarce.

In comparison with ResidualGAN, our method attains a higher visual quality of translation results. One important reason
is that ResidualGAN uses the vanilla GAN loss, which may not provide a stable gradient in its training process. Another is the residual image in ResidualGAN is not easy to learn. Instead our SG-GAN uses the objective term of WGAN-GP as the adversarial loss and adopts the proposed refined residual image learning.

Furthermore, compared with StarGAN, the proposed SG-GAN demonstrates its superior advantage in keeping the unrelated content consistency,
for example, image background. It can be interpreted that the refined residual image learning component adopted by the proposed
model can effectively help the information flow transfer and choose a good initial point to manipulate the region corresponding
to the targeted attribute.

Finally, StarGAN requires multi-attribute labels as target domains to translate a single attribute, which leads to the weakness of the model
that its translation results are affected by a set of attributes that may not be of direct interest in a specific task, especially when the training data is unbalanced.
As shown in the left of Fig.~\ref{fig:figure_archi_dis}, when gender or hair color is treated as the target attribute, skin also has been translated from slightly pale
to overly pale, in which case skin affects the translation results of other attributes.
In the right of Fig.~\ref{fig:figure_archi_dis} for every attribute, we show the absolute value of residual image between the input image and translation result to better illustrate this issue. For example, the residual images with hair color as the target for StarGAN has more highlight in facial areas. When trained on the same dataset, SG-GAN does not suffer from this problem. We explained that our special one-input-multiple-output architecture with separate training loss for every attribute could avoid this problem.

We, therefore, conclude that our proposed SG-GAN with sparsely grouped learning could be applied for facial attribute manipulation task and attains the comparable translation results at the same time.

{\bfseries Quantitative Evaluation Protocol:}

Through qualitative evaluation, the advantages of the
proposed method in attaining a high-quality image translation results and preserving
other visual attributes unrelated in the translation task are
further demonstrated. To assess the sample quality and degree of translation quantitatively,
we employed three tactics to confirm the previous conclusions.

First, following the metric proposed in~\cite{salimans2016improved},
we use the pretrained inception model to compute Inception score~(IS) on all generated images to evaluate
the maintenance of meaningful objects contained in the image
and the variety of generated images. The higher IS correspond to higher
quality generated images.

Second, we compute the classification accuracy of
translation results using the ResNet-18~\cite{he2016deep},
which is the same evaluation network used in
StarGAN~\cite{choi2017stargan}. For this previous evaluation method in StarGAN,
we notice that the classification accuracy of translation
results concerning the targeted attributes does not consider the
preservation of unrelated image attributes in a translation process. Therefore, we compute the accuracy not only on targeted attributes but also other unrelated
facial attributes to explore whether there is any side effect with accidentally modifying
these unrelated attributes in a translation process. Here we select three attributes: gender, smile and hair color in this experiment.

Additionally, to show the ability of keeping the consistency of
irrelevant content, for example, image background, we use MS-SSIM~\cite{wang2004image}
to measure the similarity in image background between translation
results and input samples. A higher MS-SSIM value corresponds to a higher
similarity between images in human perception. Specifically, we crop 10 $\times$ 10 top-left region
from both translation results and input samples as their background.

{\bfseries Quantitative evaluation:}

Table~\ref{tab:inception} shows IS on translated samples from different models. SG-GAN(ALL) obtains higher scores in most of the cases. The translated results for SG-GAN(500)
 with $500/10,0000=0.005$ labelled ratios are still high-quality and acceptable, e.g.,
 3.30$\pm$0.147 for SG-GAN(500), 3.24$\pm$0.118 for the ResidualGAN, 3.36$\pm$0.164 for StarGAN.
Overall, these scores clearly show the advantages of SG-GAN in generating
 high-quality images.

As shown in the 1st column~(Left) of Table~\ref{tab:gender_classification}, we give the classification accuracy of attribute gender on translation results with attribute gender as the target which
indicates SG-GAN achieves an acceptable degree of translation.
The accuracies of other attributes, smile and hair color, on this translation
results are shown in the 2nd and 3rd columns. In the
case of attribute smile, SG-GAN(ALL) achieves an accuracy of 86.87, higher
than both baseline models. For hair color, our model and
baseline models have similar accuracy. It indicates that our model and
StarGAN have some advantages in generated images for the maintenance of identity than ResidualGAN in general. Additionally,
for the ability keeping the consistency of irrelevant content with MS-SSIM as the metric, our model is more
capable of maintaining background unchanged, e.g., 72 for SG-GAN(All),
70 for SG-GAN(500), 64 for StarGAN and 63 for ResidualGAN, showing in the
last column. The most important is that SG-GAN(500) and SG-GAN(5000)
obtain very closed scores compared with SG-GAN(ALL) in most the cases.
The right of Table~\ref{tab:gender_classification} reports scores on hair color
as the targeted attribute for translation which could indicate a similar conclusion.

\begin{table}
\small
\begin{center}
\caption{Classification accuracy [\%] of translation results on SG-GAN(500) using Residual image learning~(RIL) and refined residual image learning~(RRIL).}
\begin{tabular}{cccccc}
\hline
Method & Gender & Smile & Hair Color \\
\hline
RIL & 18.77 & 23.67 & 21.35 & \\
Refined RIL & {\bfseries 93.26} & {\bfseries 82.67} & {\bfseries 88.91} &\\
\hline
\end{tabular}
\label{tab:ablation_study}
\end{center}
\end{table}

\begin{figure}[!htbp]
\vspace{-0.4cm}
\begin{center}
\includegraphics[width=1.0\linewidth]{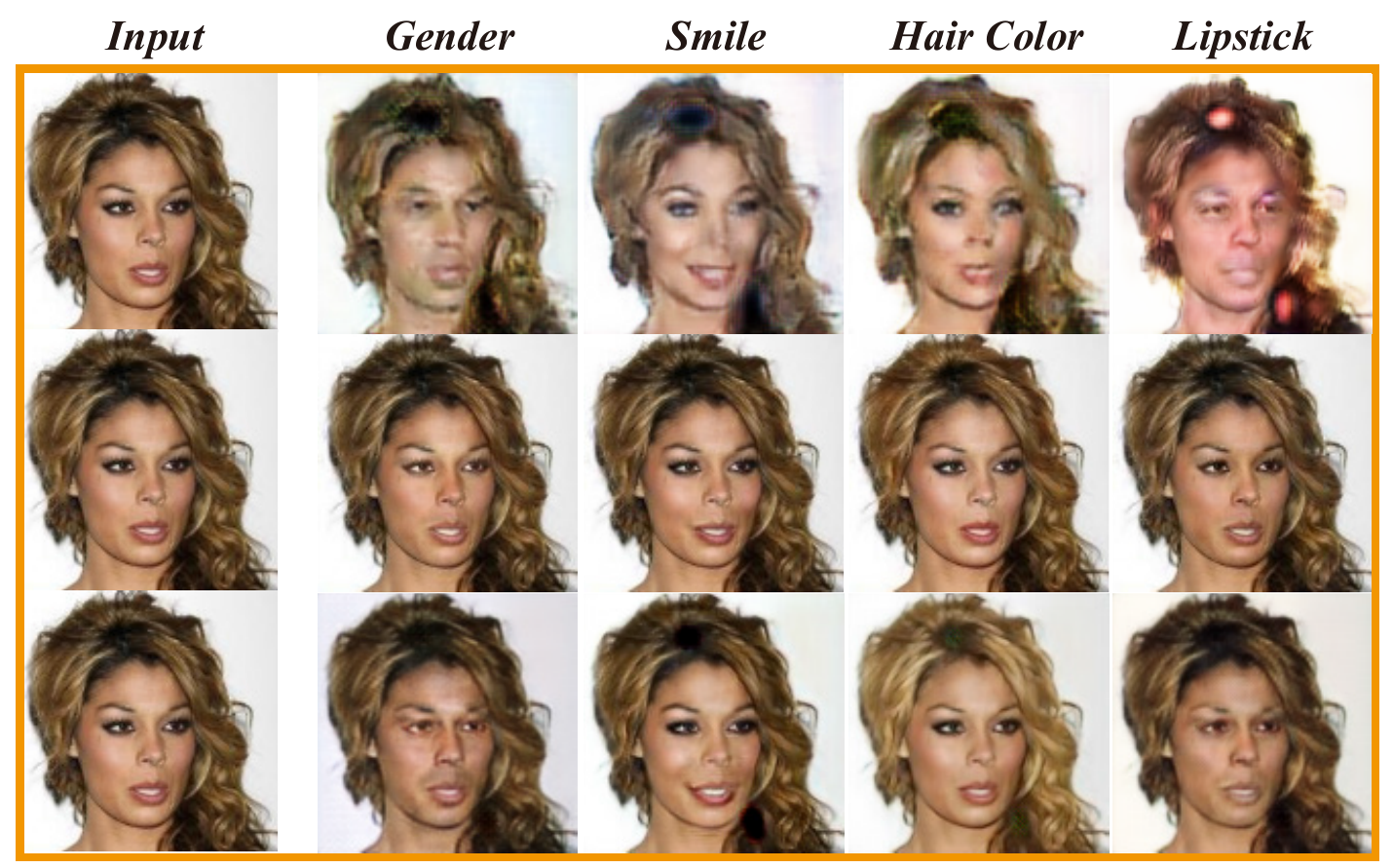}
\end{center}
\vspace{-0.4cm}
\caption{The validation of effectiveness for refined residual image learning. The three rows are the attributes translation results of SG-GAN without residual learning, SG-GAN with the original residual image learning methods, SG-GAN with the proposed refined residual image learning, respectively.}
\label{fig:ablation}
\end{figure}

\subsection{Baseline Comparison for Sparsely Grouped Learning}


As shown in Fig.~\ref{fig:comparison_unb},
we show translation results of our method and baseline methods
for pale skin on the training dataset with the different number
of the grouped-labelled dataset. The original ResidualGAN cannot be easily trained on sparsely grouped datasets because of its double-generator architecture requires input data with grouped-labels. Thus, we train ResidualGAN only using grouped data.
For StarGAN, it is also hard to be trained on sparsely grouped
datasets, because it requires original labels to compute the reconstruction loss. For a fair comparison, we modify the
reconstruction loss of StarGAN for sparsely grouped learning.
Specifically, we use the binary mask to remove the corresponding reconstruction loss of
the input image without grouped-labels and only compute the reconstruction
loss of the input $\mathbf{x}$ with grouped-labels.

For StarGAN and SG-GAN trained on the same sparsely grouped data, the translation results are shown on the second and third rows of Fig.~\ref{fig:comparison_unb}. In terms of visual quality, StarGAN's result is worse than SG-GAN trained on same data. We explain that StarGAN depends on
the reconstruction loss to preserve the identity and background content of the input image. However, their reconstruction loss requires the original grouped-labels of the input images, where it
would make no sense for the input data without labels in sparsely grouped learning. As shown in Eq.~\ref{recon_loss}, the proposed
reconstruction loss does not need the original labels of input data, which is one major advantage of the new method.

\subsection{Ablation Study}

In this section, we validate the effectiveness of the refined residual image
learning component in the proposed method by comparing the translation results
generated by SG-GAN with its several variants. As shown in Fig.~\ref{fig:ablation}, without residual image learning, SG-GAN could only generate low-quality
images that appear blurry and fail to preserve visual characteristics uninvolved in
the translation task. As shown in the 2nd and 3rd row of Fig.~\ref{fig:ablation}, in comparison with the original residual image
learning method proposed by~\cite{shen2017learning}, refined residual image learning component improves the degree of image translation. For the original residual learning, the element-wise sums between the real images and the network output would be the final translation result. However, the learned residual image tends to become sparse, and the output becomes close to the input when the adversarial loss for training has sufficient capacity.
Our refined residual image learning is to use concatenation operation of the input and feature maps instead of this element-wise sums to automatically learn their arithmetic relationship. As shown in the Table~\ref{tab:ablation_study}, quantitative evaluation further validates the effectiveness of this design.

\section{Conclusion}
We have proposed a new model, i.e. SG-GAN, to perform multiple facial
attribute manipulation with one-input multi-output architecture,
which is well suited for tackling multi-task learning and sparsely
grouped learning tasks. Tightly coupled with this learning architecture,
a refined residual image learning paradigm has been shown to
enhance the performance of the new method in image translation.
As consistently demonstrated results reported in the paper, SG-GAN
can attain comparable image translation results as multiple peer methods
and preserve unrelated region while having access to significantly fewer grouped-labels training data.

However, our results are far from uniformly positive, and SG-GAN has a few limitations, for example,
unnatural geometric translation, semantic correction, a decrease of translation quality when the grouped-labels are rare in
the training dataset. In the future, we hope to solve these problems by the proposed new model.

{\small
\bibliographystyle{ieee_fullname}
\bibliography{egbib}
}

\clearpage

\renewcommand\arraystretch{1.5}
\begin{table*}[t]
\normalsize

\section{Appendix}

\subsection{Network Architecture}\label{net_arch}
    The network architecture of SG-GAN are shown in Table~\ref{tab:Generator architevture} and Table~\ref{tab:Discrimicator architecture}. Because SG-GAN is a multi-task learning framework, it can manipulate multiple attributes at the same time, we only choose an attribute to show the architecture. Here are some notations should be noted. $n_{c}$: channel of results. $n_{t}$: range of value for the attribute. $h$: height of input images. $w$: width of input images. C: channels of images. K: size of the kernel. S: size of the stride. $P$: padding method. D: the scale of resize.

\footnotesize

\begin{center}

\begin{tabular}{cccccc}

\toprule
Part & Input Shape & Operation & Output Shape\\
\midrule
\\
encoder & $(h,w,n_{c})$ & CONV-(C64, K7$\times$7, S1$\times$1,$P_{same}$),\quad  ReLU,\quad Instance Normal &  $(h,w,64)$\\
  & $(h,w,64)$ & CONV-(C128, K4$\times$4, S2$\times$2,$P_{same}$),\quad  ReLU,\quad Instance Normal &  $(\frac{h}{2},\frac{w}{2},128)$\\
        & $(\frac{h}{2},\frac{w}{2},128)$ & CONV-(C256, K4$\times$4, S2$\times$2,$P_{same}$),\quad  ReLU,\quad Instance Normal & $(\frac{h}{4},\frac{w}{4},256)$ \\
\\
\midrule
\\
bottleneck & $(\frac{h}{4},\frac{w}{4},256)$ & Residual Block:CONV-(C256,K3$\times$3,S1$\times$1,$P_{same}$),\quad ReLU,\quad Instance Normal &  $(\frac{h}{4},\frac{w}{4},256)$ \\
           & $(\frac{h}{4},\frac{w}{4},256)$ & Residual Block:CONV-(C256,K3$\times$3,S1$\times$1,$P_{same}$),\quad ReLU,\quad Instance Normal &  $(\frac{h}{4},\frac{w}{4},256)$ \\
           & $(\frac{h}{4},\frac{w}{4},256)$ & Residual Block:CONV-(C256,K3$\times$3,S1$\times$1,$P_{same}$),\quad ReLU,\quad Instance Normal &   $(\frac{h}{4},\frac{w}{4},256)$\\
           & $(\frac{h}{4},\frac{w}{4},256)$ & Residual Block:CONV-(C256,K3$\times$3,S1$\times$1,$P_{same}$),\quad ReLU,\quad Instance Normal &   $(\frac{h}{4},\frac{w}{4},256)$\\
           & $(\frac{h}{4},\frac{w}{4},256)$ & Residual Block:CONV-(C256,K3$\times$3,S1$\times$1,$P_{same}$),\quad ReLU,\quad Instance Normal &  $(\frac{h}{4},\frac{w}{4},256)$ \\
           & $(\frac{h}{4},\frac{w}{4},256)$ & Residual Block:CONV-(C256,K3$\times$3,S1$\times$1,$P_{same}$),\quad ReLU,\quad Instance Normal &   $(\frac{h}{4},\frac{w}{4},256)$\\
\\

\midrule
\\
decoder & $(\frac{h}{4},\frac{w}{4},256)$ & DECONV-(C128,K4$\times$4,S2$\times$2,$P_{same}$),\quad ReLU,\quad Instance Normal&$(\frac{h}{2},\frac{w}{2},128)$ \\
        & $(\frac{h}{2},\frac{w}{2},128)$ & DECONV-(C64,K4$\times$4,S2$\times$2,$P_{same}$),\quad ReLU,\quad Instance Normal& $(h,w,64)$\\
        & $(h,w,64)$ & CONCAT & $(h,w,64+3)$\\
        & $(h,w,64+3)$ & CONV-(C($n_{c}$),K7$\times$7,S1$\times$1,$P_{same}$) & $(h,w,n_{c})$\\
\\
\bottomrule

\end{tabular}
\end{center}
\caption{Generator architecture}

\label{tab:Generator architevture}
\end{table*}

\renewcommand\arraystretch{1.5}
\begin{table*}[t]
\footnotesize

\begin{center}

\begin{tabular}{cccccc}

\toprule
Part & Input Shape & Operation &output \\
\midrule
discriminator & $(h,w,n_{c})$ & CONV-(C64, K5$\times$5, S2$\times$2,$P_{same}$),\quad  Leaky ReLU &  $(\frac{h}{2},\frac{w}{2},64)$ \\
              & $(\frac{h}{2},\frac{w}{2},64)$ & CONV-(C128, K5$\times$5, S2$\times$2,$P_{same}$),\quad  Leaky ReLU &  $(\frac{h}{4},\frac{w}{4},128)$\\
              & $(\frac{h}{4},\frac{w}{4},128)$ & CONV-(C256, K5$\times$5, S2$\times$2,$P_{same}$),\quad  Leaky ReLU &   $(\frac{h}{8},\frac{w}{8},256)$\\
              & $(\frac{h}{8},\frac{w}{8},256)$ & CONV-(C512, K5$\times$5, S2$\times$2,$P_{same}$),\quad  Leaky ReLU & $(\frac{h}{16},\frac{w}{16},512)$ \\
              & $(\frac{h}{16},\frac{w}{16},512)$ & CONV-(C512, K5$\times$5, S2$\times$2,$P_{same}$),\quad  Leaky ReLU &  $(\frac{h}{32},\frac{w}{32},512)$\\
              & $(\frac{h}{32},\frac{w}{32},512)$ & CONV-(C1024, K5$\times$5, S2$\times$2,$P_{same}$),\quad  Leaky ReLU &  $(\frac{h}{64},\frac{w}{64},1024)$\\

\bottomrule
$D_{cls}$ & $(\frac{h}{64},\frac{w}{64},1024)$ & CONV-(C($n_{t}$), K2$\times$2, S1$\times$1,$P_{valid}$)& $(\frac{h}{128},\frac{w}{128},n_{t})$ \\
\bottomrule
$D_{adv}$ & $(\frac{h}{64},\frac{w}{64},1024)$ & CONV-(1, K3$\times$3, S1$\times$1,$P_{same}$)&  $(\frac{h}{64},\frac{w}{64},1)$ \\

\end{tabular}
\end{center}
\caption{Discriminator architecture}

\label{tab:Discrimicator architecture}
\end{table*}

\begin{figure*}
\subsection{Additional Qualitative Results}
\vspace{-0.2cm}
\begin{center}
\includegraphics[width=0.97\linewidth]{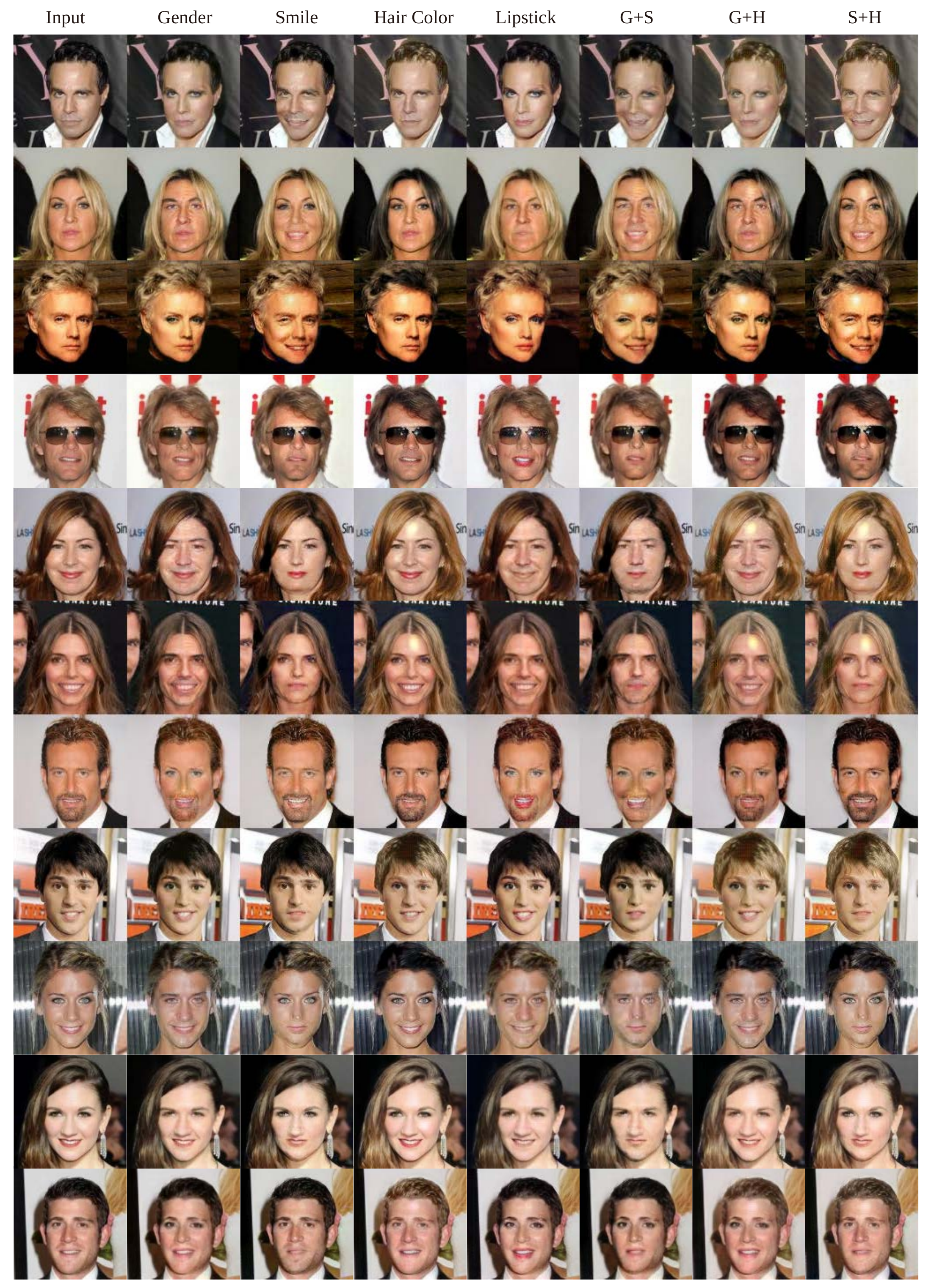}
\end{center}
\vspace{-0.4cm}
\caption{Single and multiple attribute translation results on CelebA using method SG-GAN(ALL).}
\label{fig:more_results_all}

\end{figure*}

\begin{figure*}
\vspace{-0.2cm}
\begin{center}
\includegraphics[width=0.97\linewidth]{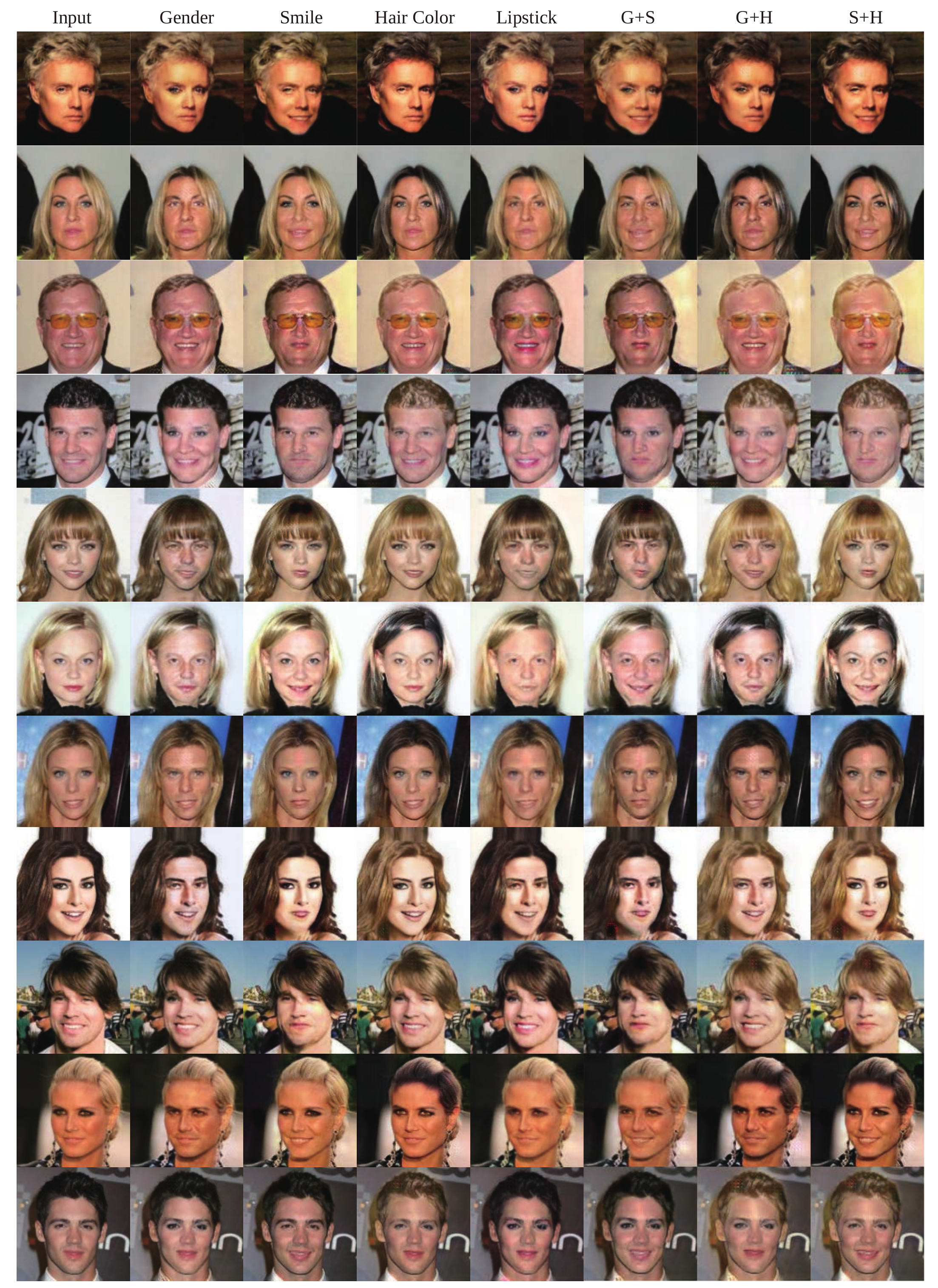}
\end{center}
\vspace{-0.2cm}
\caption{Single and multiple attribute translation results on CelebA using method SG-GAN(500).}
\label{fig:more_results_500}

\end{figure*}

\end{document}